# Robust and continuous machine learning of usage habits to adapt digital interfaces to user needs

Adaptive inference technique for interface personalization

Éric Petit, Orange Labs Grenoble, France

eric.petit@orange.com

Denis Chêne, Orange Labs Grenoble, France

denis.chene@orange.com

The paper presents a machine learning approach to design digital interfaces that can dynamically adapt to different users and usage strategies. The algorithm uses Bayesian statistics to model users' browsing behavior, focusing on their habits rather than group preferences. It is distinguished by its online incremental learning, allowing reliable predictions even with little data and in the case of a changing environment. This inference method generates a task model, providing a graphical representation of navigation with the usage statistics of the current user. The algorithm learns new tasks while preserving prior knowledge. The theoretical framework is described, and simulations show the effectiveness of the approach in stationary and non-stationary environments. In conclusion, this research paves the way for adaptive systems that improve the user experience by helping them to better navigate and act on their interface.

L'article présente une approche d'apprentissage automatique visant à concevoir des interfaces numériques capables de s'adapter dynamiquement à différents utilisateurs et stratégies d'utilisation. L'algorithme utilise des statistiques bayésiennes pour modéliser le comportement de navigation des utilisateurs, se concentrant sur leurs habitudes plutôt que sur des préférences de groupe. Il se distingue par son apprentissage incrémental en ligne, permettant des prédictions fiables même avec peu de données et dans le cas d'un environnement changeant. Cette méthode d'inférence génère un modèle de tâches, offrant une représentation graphique de la navigation avec les statistiques d'usage de l'utilisateur courant. L'algorithme apprend de nouvelles tâches tout en préservant les connaissances antérieures. Le cadre théorique est décrit et des simulations montrent l'efficacité de l'approche dans des environnements stationnaires et non stationnaires. En conclusion, cette recherche ouvre la voie à des systèmes adaptatifs améliorant l'expérience de l'utilisateur en l'aidant à mieux se repérer et agir sur son interface.

**Mots-clés additionnels :** interfaces utilisateur adaptatives, apprentissage incrémental continu, inférence bayésienne, personnalisation, modèle de tâches

**Forword**

This article was submitted in French to the IHM 2025 conference and was rejected. The reasons given include that it would be too oriented toward machine learning to speak to a community of HCI researchers and not concrete enough, as well as other reasons that we largely dispute. In light of the comments from the two reviewers, it appears that our non-parametric Bayesian approach was not understood, nor the crucial issue of "sequential, continuous and robust learning" for the design of adaptive user interfaces.

## 1 INTRODUCTION

Users are all different. Some have no particular constraints but have usage habits and preferences. Others, such as people with disabilities or seniors, may have, in addition to these habits, constraints when using a digital service. These constraints can be very diverse, of a perceptual nature (visual, auditory, tactile), of a motor nature (pointing, manipulation, speech) or cognitive (reasoning, memory, comprehension, reading...). Consequently, any service, any interface should be able to adjust to these constraints. However, habits and constraints are infinite, and it is hardly conceivable to anticipate all of them when designing a service. This is where a Machine Learning (ML) algorithm can prove highly relevant. This is what we will see with the ABIT[1] algorithm, which performs Bayesian usage statistics of the interface for each habit, each constraint specific to each individual, and consequently proposes to the user to adjust the use of the service, either by guiding them on their usual paths or by making the interface more efficient. In this scheme, the machine learning algorithm constitutes the cornerstone of the adaptive system. The latter will then be defined as a system capable of automatically modifying its characteristics according to the user's needs for a defined use in a specific context. It will notably adapt to the diversity of:

- user profiles (their preferences and physical capabilities),
- their usage habits or usage strategies (essential component of the usage context),
- the environmental contexts in which they evolve.

In this framework, we will talk about personalization of interfaces when the interface is adjusted to the specific needs of ONE single user, in reference to the taxonomy of (Simonin & Carbonell, 2007). Note that these needs may be related to sensory, actional, cognitive constraints linked to impairments, or to particular contexts (little or no light, ambient noise, sitting position, dual task, stress...). These needs may also be related to usage habits, such as the more or less frequent use of a particular service or function, or at different time scales (hour, day, week, ...). An adaptive interface should allow bringing to the user modifications of their experience to meet their specific needs by configuring the layout (Vanderdonckt, Bouzit, Calvary, & Chêne, 2020), content or functionalities of the system. This personalization may involve moving elements in an interface to reflect their priorities, highlighting their preferred interface elements and usual paths, anticipating their actions, or involving other factors related to the general design of an interface.

In the design model for adaptive interfaces (PDA-LPA) described in (Bouzit, et al., 2017), the interactional cycles evolve on both sides of the interface and feed each other. This model is based on the intersection between the Perception-Decision-Action (PDA) cycles on both the user and system sides, and the Learning-Prediction-Adaptation (LPA) cycles, on both the user and system sides. In this article, we present an adaptive inference technique aimed at modeling the usage habits of the current user. A habitual task will then be considered as a sequence of actions that the user routinely performs on the interface in order to achieve a goal (for example, making a bank transfer or launching an episode of their favorite series). This modeling of usage habits relates to the LPA axis. However, this technique does not address the other dimensions of the user profile located on the PDA axis, such as their preferences or physical capabilities (Martin-Hammond, et al., 2018) which can be addressed by other means, notably through flexible and adaptable interfaces in a universal design logic (Stephanidis, 2001). We also do not consider in this article the adaptation of the interface to the environment or to the interaction device as developed by *plasticity* (Thevenin & Coutaz, 1999) and which can be treated separately. We also differentiate ourselves from recommendation systems that are generally associated with taking into account the preferences of a group of users and that involve machine learning techniques requiring prior training of the model on offline data, as in (Verma, Patel, & Abhari, 2009). We also do not address software architecture models, although the design of adaptive interfaces generally requires appropriate software engineering, particularly *model-driven* (Akiki, Bandara, & Yu, 2014). It should be noted that in these architecture models, adaptation rules are generally fixed and activated based on the characteristics of the usage context, declared or inferred, such as the user's age or the estimation of ambient light. But, as (Giuffrida, Céret, Dupuy-Chessa, & Poli, 2019) pointed out, these rules rarely take into account the uncertainty about the perception of the situation. This is why the authors propose using fuzzy logic to flexibly and dynamically manage the combination of adaptation rules. However, this approach does not integrate the user's habits or their usage strategies.

---

[1] Adaptive Bayesian Inference Technique - É. Petit - ORANGE
IDDN FR 001 230021 002 S P 2020 000 20700

It should be noted that research on interface personalization (Oppermann, 1994) (McGrenere, Baecker, & Booth, 2002) (Bunt & Conati, 2004) has shown that users often fail to configure/personalize their interface themselves, and that assistance provided by the system would be desirable, especially since user habits change over time. This is the approach we adopt, combined with the "mixed-initiative" approach advocated by Horvitz (Horvitz, 1999). In this approach, the decision relies both on the system's ability to infer probabilities about the user's intentions, and on the continuous evaluation of the benefit/cost ratio of system-initiated adaptations. In addition, the system can engage in a dialogue with the user, particularly in the case of an ambiguous situation.

In (Gajos, Czerwinski, Tan, & Weld, 2006), user interactions are taken into account over time to optimize editing tasks within Microsoft Word, highlighting the benefit of an adaptive approach to "split menus" (Sears & Shneiderman, 1994). In this study, the authors point out the importance of the benefit/cost (human) ratio of adaptations in the acceptance of these mechanisms, as well as the criteria of stability, reliability (in the sense of "accuracy") and predictability of adaptation (Gajos, Everitt, Tan, Czerwinski, & Weld, 2008). The reliability of predictions combined with the predictability of the interface are a key component of human trust in the machine, which trust plays a crucial role in mediating interactions between humans and automated systems as discussed in (Lee & See, 2004). For a state of the art of concepts to manipulate in interface adaptation, see (Bouzit, 2017). Note that the prototype of (Gajos, Czerwinski, Tan, & Weld, 2006) was based on two very simple inference algorithms, one focused on calculating the frequency of use of functions, and the other on calculating the recency of use. It therefore did not allow modeling a very rich usage context, such as taking into account a sequence of actions. In fact, this type of approach seems restricted to simple interface adaptations.

We aim here at a more general personalization paradigm that would allow the user to better navigate their interface and act more easily according to their own usage habits. This paradigm focuses on how the user acts on their interface in order to optimize their interaction. It has been explored for specific applications, such as news browsing on a smartphone (Constantinides & Dowell, 2018) to facilitate navigation and go beyond content recommendation. It is based on an interaction model designed as an extension of the *user profile*. Its generation involves, on the one hand, continuous capture of user activity (commands, selections, context, etc.) and, on the other hand, heuristics or a predictive model capable of inferring the characteristics of the *user profile* over time. In (Constantinides & Dowell, 2018), the inference of the three so-called "high-level" characteristics goes through a classifier whose supervised learning was done offline based on a corpus of activity logs. However, an operational implementation would require automating the generation of the user profile and making it evolve over time, so that it continuously adjusts to the changing behavior of the user. This remains a difficult problem to date. With this general framework established, we focus on the question of how an application can automatically create and continuously update an interaction model specific to the current user?

To answer this, we start from the principle that the predictive model must satisfy several properties, particularly in terms of robust inference and continuous learning, which is not without difficulties. Indeed, in machine learning, predictive models are generally static, resulting from a two-phase learning process (Wikipédia, Apprentissage supervisé, 2025). They cannot adapt over time (Shaheen, Hanif, Hasan, & Shafique, 2022) or only marginally through "fine tuning" techniques (IBM, 2025). In fact, they regularly need to be retrained. But in a dynamic environment that changes regularly, such as human behavior, this approach is inappropriate. This is why the emerging field of "Continual Learning," also called "incremental learning" or "lifelong learning" (Chen & Liu, 2018) (Luo, Yin, Bai, & Mao, 2020) (Shaheen, Hanif, Hasan, & Shafique, 2022) (Hoi, Sahoo, Lu, & Zhao, 2021) studies algorithms capable of learning continuously in a changing environment, with the ability to retain and accumulate past knowledge, to be able to draw inferences from it, to use this knowledge to learn new ones more easily and thus solve new tasks. More precisely, we place ourselves here in the case of autonomous systems learning from a continuous data stream (Shaheen, Hanif, Hasan, & Shafique, 2022). We particularly aim for self-supervised incremental online learning. This means that only one example is presented at a time to update the predictive model (Hoi, Sahoo, Lu, & Zhao, 2021). Such systems therefore continue to learn after their deployment autonomously and economically. However, this new field remains very oriented towards "neural networks," which implies learning data in large numbers, and which makes it difficult to obtain reliable uncertainty measures on predictions (Cornuéjols, Miclet, & Barra, 2018). On the one hand, we do not have large amounts of data, since the capture focuses on the activity of a single user, and on the other hand, this measure of uncertainty (or degree of confidence) is central to being able to manage at any time the level of automation of the service and the dialogue with the user, as Horvitz explains (Horvitz, 1999). The latter specifies that in the presence of uncertainty, one should prefer to "do less" but correctly. This is why we advocate a probabilistic modeling, as it is the only one capable of quantifying and managing uncertainty in a rigorous manner.

In part 2, we discuss the issue of artificial learning modeling. In part 3, we propose a Bayesian theoretical framework to overcome these limitations with the implementation of an original adaptive inference technique. In addition to its qualities of robustness and continuous learning, this technique also offers the advantage of being able to extract at any time a task model from the activity of the current user. Finally, in parts 4 and 5, we show, through simulations of navigation tasks within a hierarchical menu, the validity of this approach in the stationary case (constant habits) and in the non-stationary case when usage habits evolve over time.

## 2   THE IMPORTANCE OF MODELING

In this UX framework, the modeling problem aims to allow the machine to interpret the user's actions in order to deduce intentions and trigger interface adaptations. This problem, reduced to modeling the user's activity, involves a limited number of variables, on the one hand: explanatory variables, themselves of two kinds: context variables (place, time, etc.) and action variables (selections of buttons, links, menus, various options in the interface) and on the other hand dependent variables that we seek to predict, such as the sequence of actions to reach the final state of the task. As we mentioned, this modeling involves the design and implementation of a machine learning algorithm with specific characteristics, namely, capabilities to:

- Provide predictions with a reliable confidence measure allowing decision-making.

- Learn quickly from little data by allowing decision-making as early as possible.
- Take into account an environment that changes regularly.

While AI has seen spectacular progress in recent years with the arrival of foundation models (Wikipédia, Modèle de fondation, 2025) (Cornuéjols, Miclet, & Barra, 2018), it still remains underperforming with regard to these three characteristics. We will see in the rest of this article, how in a "small data" framework it is possible to propose a solution compatible with these properties.

Let's start by formalizing the problem as follows: let's say $X_1, X_2, A_1, A_2$ are the explanatory variables, where $X_1$ and $X_2$ are two context variables (such as "day" and "place") and $A_1, A_2$ two action variables (selections of buttons, links, menus, etc.) corresponding for example to a beginning of interaction with the machine for a given application. Let $A_3, A_4, A_5$ be three dependent variables (other selections of buttons, menus, etc.) corresponding to the user's future actions to accomplish their task. To fix ideas, the task may correspond, for example, to a bank transfer or the launch of a TV program. These variables are of attribute-value type and have a finite number of values. A particular realization of the task then corresponds to a sequence of values, for example the sequence: $\{x_{1i}, x_{2i}, a_{1i}, a_{2i}, a_{3i}, a_{4i}, a_{5i}\}$ where $a_{5i}$ is the final state of the task. The modeling problem boils down to predicting the sequence $\{a_{3i}, a_{4i}, a_{5i}\}$ given the current sequence $\{x_{1i}, x_{2i}, a_{1i}, a_{2i}\}$. This is in fact a multi-label classification problem for which the optimal mathematical solution is provided by Bayesian probability theory. Let us recall that this theory allows to formalize a mode of rational reasoning in the presence of uncertainty. It allows to go back to the "probability of causes" according to Pierre Simon Laplace, that is, to the *posterior* probability. Thus, the search for the exact solution involves calculating the plausibility of each hypothetical sequence $\{a_{3i}, a_{4i}, a_{5i}\}$, given the known sequence $\{x_{1i}, x_{2i}, a_{1i}, a_{2i}\}$, i.e., the probability: $P(a_{3i}, a_{4i}, a_{5i} / x_{1i}, x_{2i}, a_{1i}, a_{2i})$. How then to carry out this calculation and why is a probabilistic approach required rather than a "deep learning" approach based on neural networks?

To introduce Bayesian inference and its modern interpretation since the Cox-Jaynes theorem (Wikipédia, Théorème de Cox-Jaynes, 2025), we start from 2 logical propositions $A$ and $B$ with truth value True or False and their respective negations $\bar{A}$ and $\bar{B}$. Suppose that "$A$ is the cause of $B$" and that "$B$ is the proof of $A$" (the explanatory variable). Suppose that $B$ is 99% effective in terms of *likelihood*, precisely:

- If $A$ is true, then $B$ is true with a probability of 99%, i.e., $P(B|A) = 0,99$ ;
- And if $A$ is false, then $B$ is false with a probability of 99%, i.e., $P(\bar{B}|\bar{A}) = 0,99$.

It is understood that it is $B$ that we observe to try to predict $A$. So, the question asked is the following: If proposition $B$ turns out to be true, what can we say about $A$ ? In reality, we cannot answer this question, even if intuition might lead us to think that $A$ is true (at least plausible). We are in fact missing the information of the degree of knowledge of $A$ in the form of its *prior probability*, a key concept of Bayesian theory, denoted $P(A)$. Let's then consider for $A$ a very low *prior probability*, namely: $P(A) = 0,001$. In this case, we can approximate Bayes' formula by the ratio: $P(A)/P(B|\bar{A})$ with $P(B|\bar{A}) = 0,01$ (probability of false positives). It follows that the *posterior probability* of $A$ given $B$ is 10%, actually reflecting a low confidence in hypothesis $A$. Thus, despite an apparently very high proof of $A$ (*likelihood* of $B$ of 99%), Bayesian inference does not conclude to its plausibility. Let us recall that in the Bayesian interpretation, probability measures a degree of state of knowledge about a hypothesis. Here the information provided by $B$ is not strong enough to change the conclusion, but it allows to strongly revise the initial belief in proposition *A*, increasing its probability from 0.1% to 10%. Note that the impact of the *likelihood* of *B* on the *posterior probability* of *A* is all the stronger as the *prior probability* of *A* is low.

Let us remember from this example that the *prior probability* is a fundamental component of Bayesian statistical inference. However, in the case of parametric models, especially neural networks, it seems difficult to take it into account in a rigorous way, which generally leads to systems that do not know how to doubt reasonably. Indeed, the classical ML approach consists in constructing, through supervised learning using a training database *data*, the best predictive model $w^*$. But probability theory teaches us on the contrary that the optimal prediction is actually given by a weighted average of predictions from a set of predictive models. More precisely, for a given input $\bar{x}$ each prediction component $P(y|\bar{x},w)$ must be weighted by the *posterior probability* of its model $w$, i.e. $P(w|data)$. But in practice, ML model training often consists in retaining only one term in the calculation of the average, the one corresponding to the best parametric model in the sense of the *maximum a posteriori* (MAP criterion), or even in the sense of the *maximum likelihood*. This type of approximation makes it difficult to reliably estimate a degree of uncertainty on predictions.

As mentioned in the introduction, another important limitation of "deep learning" is its inability to adapt to an environment that changes regularly over time. Indeed, this type of algorithm is subject to an instability phenomenon known as "catastrophic forgetting" (Luo, Yin, Bai, & Mao, 2020) which appears when one tries to learn a new task without replaying the past examples that were used to learn the old tasks. This results in a degradation of the model's knowledge related to the old tasks. This problem manifests itself when the statistical characteristics of the data evolve over time ("data drift"). It is intrinsic to the backpropagation algorithm used in the minimization of the cost function (Cornuéjols, Miclet, & Barra, 2018). Several strategies have been developed to counter the problem of "catastrophic forgetting", but they do not seem to be fully developed to date. More generally, few ML algorithms to date have the capability of sequential learning consisting of being able to learn tasks (or classes), one after the other, as they appear in the data. This type of learning also implies a mechanism of progressive forgetting of tasks that have become obsolete, i.e., a form of unlearning of the model, while ensuring the plasticity/stability trade-off (Luo, Yin, Bai, & Mao, 2020) (Chen & Liu, 2018).

Finally, the question of deploying the predictive model (putting it into production) is crucial in the case of adaptive interfaces. Indeed, the classical ML approach, which consists in training a predictive model on offline data, validating it on another dataset and then putting it into production, is not appropriate here where the model needs to be constantly updated to follow the evolution of changing human behavior. It is also desirable that the system can make decisions as early as possible, that is, from the moment the user starts using the interface, in other words, without having to wait for a long period of use. Because the objective is to be able to offer the user adaptations from the first uses, in particular to facilitate access to their usual paths and functions. Consequently, it is important to have at all times a "best" predictive model capable of producing robust inferences "as early as possible". This property of "continuous deployment" remains a hard point of machine learning.

To conclude, the classical ML approach is posed as a mathematical optimization problem, but this does not offer all the desirable properties for a UX application. This is why we propose a non-parametric Bayesian approach capable of handling non-stationary data and producing causal hypotheses related to the user's intentions and associated with real probabilities, the latter constituting reliable indicators of confidence. Therefore, the degree of confidence of a prediction will determine the type of adaptation to trigger at the interface level, as discussed in (Horvitz, 1999) (Petit & Chêne, 2021). We present our theoretical approach below.

## 3   OPTIMAL AND ADAPTIVE BAYESIAN LEARNING

Given that human behavior is constantly changing, how can we design a robust machine learning technique to continuously model user habits with the aim of improving their experience? It became clear to us that to address this question, an optimal mathematical approach was necessary. Indeed, the goal here is not simply to produce predictions but also to be able to "weigh" them accurately so that the system can make autonomous decisions. The entire challenge is to be able to calculate unbiased confidence degrees on all predictions or causal hypotheses. In part 2, we saw that the basic problem comes down to predicting the sequence of actions related to an ongoing user task. We formalized it as follows: given a current sequence $\{x_{1i}, x_{2i}, a_{1i}, a_{2i}\}$ how can we predict the likely sequence of actions $\{a_{3i}, a_{4i}, a_{5i}\}$ ? Or more formally, how can we infer its *posterior probability*: $P(a_{3i}, a_{4i}, a_{5i}|x_{1i}, x_{2i}, a_{1i}, a_{2i})$ ? To achieve this, we first solve the single-label classification problem consisting of predicting the value of a single action output $Y$ (for example $a_{3i}$) from a set of premises $\{x_1, x_2, x_3, x_4, ...\}$. The optimal Bayesian network is in this case represented by the graph in Figure 1. This structure describes the causal dependency relationships between the discrete random variables involved, here considering the variable $Y$ as the causal hypothesis (e.g., the dependent variable $A_3$ in the action sequence example). In Bayesian logic, we seek to trace back to the causes, here the user's intention represented by $Y$, given the previous actions or more generally the usage context (the $X_i$).

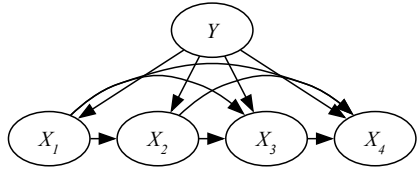

Figure 1 : Optimal bayesian network for single-label classification

This representation allows us to express the *joint probability* of all variables in the form of a product of probabilities, namely:

$$P_{X_1,X_2,X_3,X_4,Y}(x_1, x_2, x_3, x_4, y) = P_{X_1|Y}(x_1|y) * P_{X_2|X_1,Y}(x_2|x_1, y) * P_{X_3|X_2,X_1,Y}(x_3|x_2, x_1, y) * P_{X_4|X_3,X_2,X_1,Y}(x_4|x_3, x_2, x_1, y) * P_Y(y) \quad (1)$$

Using Bayes' formula, we deduce the *posterior probability* of $Y$, namely:

$$P_{Y|X_1,X_2,X_3,X_4}(y|x_1, x_2, x_3, x_4) = \frac{P_{X_1,X_2,X_3,X_4,Y}(x_1,x_2,x_3,x_4,y)}{P_{X_1,X_2,X_3,X_4}(x_1,x_2,x_3,x_4)} \quad (2)$$

Based on these formulas, we have developed an algorithm, named ABIT, that adaptively estimates the various probabilities involved. ABIT also uses digital filtering theory to construct adaptive estimators of relative frequencies. These are built from an infinite impulse response (IIR) filter whose time constant can be controlled. Indeed, one of the difficulties of the problem is that conditional probabilities cannot be updated at the same rate, particularly because of their dependence on parameters whose occurrences are governed by distinct probability laws. Therefore, to ensure a coherent combination of probabilities in equations (1) and (2), each adaptive estimator is equipped with a mechanism to adjust its integration memory in real time.

Furthermore, during the model initialization phase, we choose to assign low *prior probabilities* to output hypotheses when they first appear, knowing that we cannot know in advance the number of outcomes. This initialization strategy is advantageous for learning and generalization. Indeed, for a given random variable, low probabilities create "available space" in the probability distribution, thus leaving room for new states that may appear later. And as mentioned in §2, when the *prior probability* is small, the system remains more sensitive to the *likelihood* of the observed data; in other words, new observations will have a significant impact on the *posterior* distribution. This initialization choice therefore offers greater flexibility to allow the model to absorb new hypotheses, in accordance with the epistemic interpretation of Bayesian probability stating that weak knowledge about a hypothesis should be associated with a low probability.

In practice, ABIT dynamically creates the Bayesian network from the user's activity data. This network is initially empty and builds progressively over time. More formally, let $\{x_{1t}, x_{2t}, x_{3t}, \cdots, x_{N_t} | y_t\}$ denote the current data associated with time 't', or with vector notation: $(\bar{x}_t, y_t)$. Then consider a data stream of the form $(\bar{x}_1, y_1)$, $(\bar{x}_2, y_2),\ldots,(\bar{x}_t, y_t)$. The algorithm then successively creates incrementally a sequence of predictive models: $M_1$, $M_2,\ldots, M_t$. More precisely, from the current model $M_{t-1}$ and a new data point $(\bar{x}_t, y_t)$, the algorithm generates the new model according to a recurrence relation of the form: $M_t = F(M_{t-1}, (\bar{x}_t, y_t))$. The latter corresponds intuitively to a small modification of the previous model $M_{t-1}$. Note that only the current data $(\bar{x}_t, y_t)$ is used to update the Bayesian network and without this data being stored in memory. In this way, ABIT performs iterative incremental learning allowing it to follow the evolution of the data, while smoothing the computational load.

In this adaptive probabilistic scheme, no prior training of the network is required. Moreover, the predictive model is at all times "at its best" considering past interactions. The inference phase can therefore begin at the same time as the learning phase, as soon as the first data point is received. In practice, a configuration parameter allows setting the memory depth of the algorithm. It depends on the complexity of the input data and their variability. This hyperparameter is similar to an analysis window width.

### 3.1 Sequential modeling

We now solve the problem of predicting action sequences. This involves a *multi-label classification* approach, which led us to generalize the ABIT algorithm. This new algorithm, named ABIT-H (H for hierarchy), thus applies to a sequence of *tokens* (actions or usage contexts) forming a sequence of any length where each *token* of rank *i* is associated with a discrete random variable $W_i$ of categorical or nominal attribute type. We generally assume that each *token* of rank *i* is statistically dependent on the previous and following *tokens*. Without loss of generality, consider a sequence of length 7 denoted $\bar{s} = \{w_1, w_2, w_3, w_4, w_5, w_6, w_7\}$. Suppose that the beginning of the sequence $\bar{q} = \{w_1, w_2, w_3\}$ is known and that we want to predict the end of the sequence, namely: $\bar{r} = \{w_4, w_5, w_6, w_7\}$. The conditional joint probability of $\bar{r}$ given $\bar{q}$ is expressed, according to the chain rule of conditional probabilities, as follows:

$$P(w_4, w_5, w_6, w_7|\bar{q}) = P(w_4|\bar{q}) \times P(w_5|w_4, \bar{q}) \times P(w_6|w_5, w_4, \bar{q}) \times P(w_7|w_6, w_5, w_4, \bar{q}) \qquad (3)$$

The sequence $\bar{q}$ is here considered as the "prompt". This prompt typically corresponds to the sequence of actions already performed by the user during an interaction task. The evaluation of any sequence $\bar{r}$ then involves calculating expression (3). To do this, ABIT-H automatically instantiates an instance of ABIT for each *token* rank in the sequence being considered. Thus in (3), the instance associated with rank 4 will be responsible for calculating the probability $P(w_4/\bar{q})$, the one associated with rank 5 for calculating $P(w_5/w_4, \bar{q})$, and so on up to rank 7. In total, 7 instances of ABIT will be needed to model the entire sequence, including the prompt. Note that these instances are created dynamically during the user's interaction with their service and that they are shared by all sequences.

It follows that the total number of instances will be equal to the length of the longest observed sequence, denoted $L_{max}$. The global predictive model is then represented by a hierarchy of $L_{max}$ Bayesian networks. Finally, for a given prompt, the most probable predicted sequence is obtained by applying the maximum a posteriori criterion, evaluating all possible sequences $\bar{r}$.

It should be noted that, so far, no approximations have been made, in other words, the calculated conditional joint probability corresponds to the optimal normative solution. However, the latter can lead to a significant number of calculations in the case of long sequences and if the data has high entropy. Therefore, we have introduced a second configuration parameter, denoted $k$, to limit the correlation depth of the variables and consequently the complexity of the model. In this controlled approximation, it is assumed that each *token* of rank $n$ is statistically dependent on the previous *tokens* up to rank $(n-k)$. The hyperparameter $k$ will be called the "Markov order". In particular, for order 1 ($k=1$) each *token* depends only on the previous *token* $w_{n-1}$ (classical first-order Markovian hypothesis). The conditional joint probability is then expressed by:

$$P(w_7, w_6, w_5, w_4|\bar{q}) = P(w_7|w_6) \times P(w_6|w_7) \times P(w_5|w_4) \times P(w_4|w_3)$$

For order 2, the expression of the joint probability becomes:

$$P(w_7, w_6, w_5, w_4|\bar{q}) = P(w_7|w_6, w_5) \times P(w_6|w_5, w_4) \times P(w_5|w_4, w_3) \times P(w_4|w_3, w_2)$$

The hyperparameter $k$ thus sets the correlation depth of the predictive model. It allows finding the best compromise between robustness and complexity. A high value of $k$ allows the algorithm to capture statistical dependencies between *tokens* that are far apart within a sequence. Under this hypothesis where the value of $k$ is sufficiently high, the model will be able to faithfully extract the logical structure of the data. This is what we will see in the next section which deals with the validation of our algorithm, and which also introduces the *task model*.

## 4   VALIDATION OF THE ABIT-H APPROACH IN THE STATIONARY CASE

We present a validation of the ABIT-H algorithm in the stationary case using a simulation of navigation paths within a complex hierarchical menu, as illustrated in Figure 2. Indeed, unlike a simple tree structure (linear hierarchical menu), the menu structure has nodes with several incoming paths (those underlined), for different levels of depth. This choice of a complex structure is motivated by the desire to cover all types of menus, whether linear or not[2].

From this menu structure, we selected a set of 10 representative paths (or sequences) of a user's supposed usage (cf. Table 1), each path being assimilated to a particular task consisting of validating a series of sub-menus. Some paths use the same nodes, for example the 3 paths numbered 2, 3, and 6 in Table 1 pass through node "3b" and continue through node "4b" or "4c". This means that knowledge of a node (here "3b") is not necessarily sufficient to predict its outcome without ambiguity. In this specific case, one must additionally know the path of arrival to node "3b", for example "1a 2a" for path No. 2. It therefore appears that a Markovian modeling of order higher than

---

[2] Note that non-linear structures are hidden in basic behaviors, as in the example of a user at home having two usage habits represented in the form of sequences: morning → kitchen → "France-Inter"; evening → kitchen → "Radio Nova", modeled by 2 explanatory variables and one conclusion variable (the media listened to). The user's behavior is then determined by the first condition (the time of day), the second condition (the place) being common to both behaviors.

1 is required here to remove any ambiguity. For this simulation, we choose the optimal configuration of ABIT-H for which the *Markov order* is not constrained.

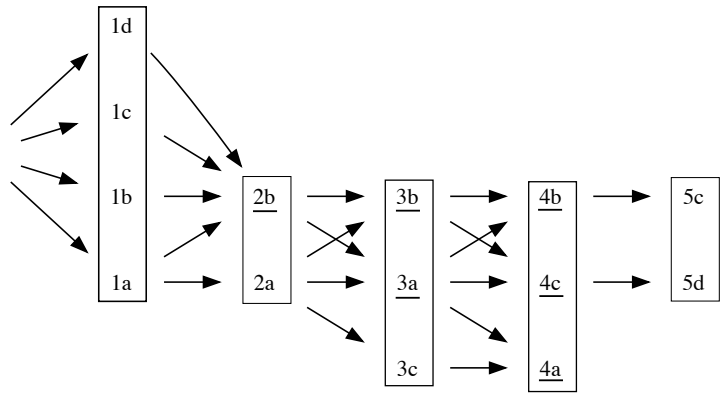

Figure 2 : Hierarchical menu structure

The simulation consists of generating the training data by random sampling from this set of sequences, having previously duplicated some of them (see the 'copies' column in Table 1). The set therefore contains 13 sequences, and its probability distribution is non-uniform due to the presence of copies. Under these controlled conditions, it is possible to theoretically calculate the joint probability of each path (see Table 1) which will be expressed on the logarithmic scale of *evidence* in *decibans* (dB) (Wikipédia, Notation d'évidence, 2025).

Table 1 : Valid paths of the simulation with their number of copies and their theoretical *evidence*

| N° | Sequences | Copies | Evidence (dB) |
|---|---|---|---|
| 1 | "1a 2a 3b 4b" | ×1 | -11 |
| 2 | "1a 2a 3b 4c" | ×3 | -5 |
| 3 | "1a 2b 3b 4b" | ×1 | -11 |
| 4 | "1a 2a 3a 4a" | ×2 | -7 |
| 5 | "1a 2a 3c 4a" | ×1 | -11 |
| 6 | "1b 2b 3b 4b" | ×1 | -11 |
| 7 | "1c 2b 3a" | ×1 | NaN |
| 8 | "1c 2b 3a 4b" | ×1 | NaN |
| 9 | "1c 2b 3a 4b 5c" | ×1 | -5 |
| 10 | "1d 2b 3a 4c 5d" | ×1 | -11 |

This choice of scale is motivated by the objective of making the interpretation of probabilities more intelligible: the degree of belief then resembles an accumulation of positive or negative *evidence*. This scale connects with cognitive sciences, knowing that a variation of *evidence* of 1dB would be the smallest variation of belief perceptible by a human, according to neuroscientist Stanislas Dehaene (Dehaene, 2025). The *evidence* in decibans is defined by the following formula:

$$Ev(p) = 10 \, log_{10}\left(\frac{p}{1-p}\right)$$

To set the ideas, 0 dB represents a probability of 50%; 20 dB a ratio of 100 to 1 (99%) and −20 dB a ratio of 1 to 100 (1%).

Regarding the theoretical calculation of the joint probability, only complete paths are considered. This therefore excludes partial paths No. 7 and No. 8. For example, for sequence No. 2, we can write:

$$P(1a, 2a, 3b, 4c) = P(4c|1a, 2a, 3b) \times P(4c) = \frac{3}{4} \times \frac{(1+3)}{13} = 0{,}23$$

This gives an *evidence* of −5.23 dB, rounded to −5 dB.

Furthermore, in order to obtain high precision on the iterative calculation of the probability distributions of the Bayesian networks, we set the number of random draws to 3900, or an average of 300 occurrences per sequence (300×13). In addition, we chose a sufficiently wide analysis window, set at 200, to allow for good statistical integration of the information. More precisely, the protocol consists of going through the set of 13 sequences 300 times, taking care each time to shuffle it randomly. Each sequence obtained is then an input data that feeds the algorithm, the latter performing incremental learning of the predictive model. At the end of the test, the model is queried with an empty prompt to infer entire sequences from the origin. The result is a list of 8 predicted sequences ranked in descending order of joint probabilities. The 3 most probable sequences are in order as follows (see Table 2):

Table 2 : Ranked most probable sequences at the end of the simulation

| 1 | 1a(0.62) 2a(0.87) 3b(0.57) 4c(0.75) —> (−5 dB) | seq. n°2 |
|---|---|---|
| 2 | 1c(0.23) 2b(1.0) 3a(1.0) 4b(1.0) 5c(0.97) —> (−5 dB) | seq. n°9 |
| 3 | 1a(0.62) 2a(0.87) 3a(0.29) 4a(1.0) —> (−7 dB) | seq. n°4 |

The conditional transition probabilities are indicated in parentheses after each *token* on a classical scale between 0 and 1. The joint probability of the global sequence is expressed in *decibans* and is displayed at the end of the line. It is obtained by multiplying the conditional probabilities along the considered path, then converting it to *evidence*. We note that the values obtained for these 3 best paths correspond well to the expected theoretical values (see Table 1), namely, −5 dB, −5 dB et −7 dB for sequences No. 2, No. 9 and No. 4, respectively. We conclude that the algorithm converges well to the optimal mathematical solution.

Regarding the predictive model, it consists of 5 Bayesian networks, one per level of hierarchical depth. Its global size is 165 parameters, reflecting the degree of complexity of the data, for a maximum *Markov order* of 4. Note that given the structure of the data, a *Markov order* of 3 would *a priori* be sufficient for modeling. We verify this by replaying the simulation and observing the same results, except for the size of the predictive model reduced to 155 parameters.

### 4.1 Case of a suboptimal modeling

In order to test the influence of the *Markov order* on the prediction, we replayed the simulation with a suboptimal order of 2. The result of the inference is as follows (see Table 3):

Table 3 : Ranked most probable sequences at the end of the simulation with a *Markov order* of 2

| 1 | 1a(0.62) 2a(0.87) 3b(0.57) 4c(0.75) —> (-5 dB) | seq. n°2 |
|---|---|---|
| 2 | 1a(0.62) 2a(0.87) 3a(0.28) 4a(1.0) —> (-7 dB) | seq. n°4 |
| 3 | 1c(0.23) 2b(1.0) 3a(1.0) 4b(0.67) 5c(0.98) —> (-8 dB) | seq. n°9 |
| 4 | 1a(0.62) 2a(0.87) 3b(0.57) 4b(0.25) —> (-11 dB) | seq. n°1 |
| 5 | 1a(0.62) 2a(0.87) 3c(0.14) 4a(1.0) —> (-11 dB) | seq. n°5 |
| 6 | 1a(0.62) 2b(0.12) 3b(1.0) 4b(1.0) —> (-11 dB) | seq. n°3 |
| 7 | 1c(0.23) 2b(1.0) 3a(1.0) 4c(0.33) 5d(0.98) —> (-11 dB) | ? |
| 8 | 1b(0.08) 2b(1.0) 3b(1.0) 4b(1.0) —> (-11 dB) | seq. n°6 |
| 9 | 1d(0.08) 2b(1.0) 3a(1.0) 4b(0.67) 5c(0.98) —> (-13 dB) | ? |
| 10 | 1d(0.08) 2b(1.0) 3a(1.0) 4c(0.33) 5d(0.98) —> (-16 dB) | seq. n°10 |

The model size is then only 122 parameters, but at the cost of a loss of performance resulting in probability estimation errors (as for sequence No. 9) and the appearance of invalid sequences of the "hallucination" type, such as the sequence "1d, 2b, 3a, 4b, 5c" (rank 9) coming from the recombination of the two sequences No. 9 and No. 10. This demonstrates that a model that is too simple (too permissive) compared to the input data will be subject to hallucinations, hence the importance of being able to control its complexity and aim for the optimal solution.

### 4.2 Towards a task model

This Bayesian approach therefore makes it possible to precisely model sequences of events. In our UX framework, this will involve modeling sequences of actions performed by the user when carrying out a task within a digital interface. From there, we show that it is possible to generate a *task model* concentrating the knowledge of the user's interaction activity and capable of highlighting their usage habits as well as the logical structure of navigation. To do this, we have developed a function that allows the algorithm to extract a graphical representation of this model, based on the inferences produced, for a given prompt. This *task model* is therefore generated by the algorithm by querying its predictive model. It uses the Dot language (GraphViz, 2025) to describe the logical structure of the data in the form of a graph, incorporating usage statistics. This description in the form of a script is saved in a text file. Its display involves conversion to vector format based on the Graphviz tool (GraphViz, 2025). For our simulation, we obtain the *task model* in Figure 3.

This representation highlights the sequences of actions of the user accompanied by their conditional and joint probabilities (at the end of the path and in *decibans*). It then becomes easy to identify the user's usage strategies, particularly their most usual paths, indicated on the graph using numbers (1), (2) and (3) or colors: red, green and blue, respectively. These 3 highlighted paths naturally correspond to the most probable predicted sequences, namely sequences No. 2, No. 9 and No. 4. Furthermore, the graph allows understanding the logical structure of the interface and identifying in particular the menus that are shared between several tasks, such as the menu "3a" which is common to sequences No. 9 and No. 4. In the present case, we can see that the structure of the *task model* faithfully reproduces the hierarchical structure of the menu. Note that in a human-machine interaction situation, the algorithm will be able to extract the logical structure of the application that corresponds to the user's actual activity. The reconstructed structure may therefore be partial if the user has a restricted use of their application, i.e., centered on certain tasks. In general, the *task model* will reflect the user's level of mastery and the adequacy of the functional complexity to their needs.

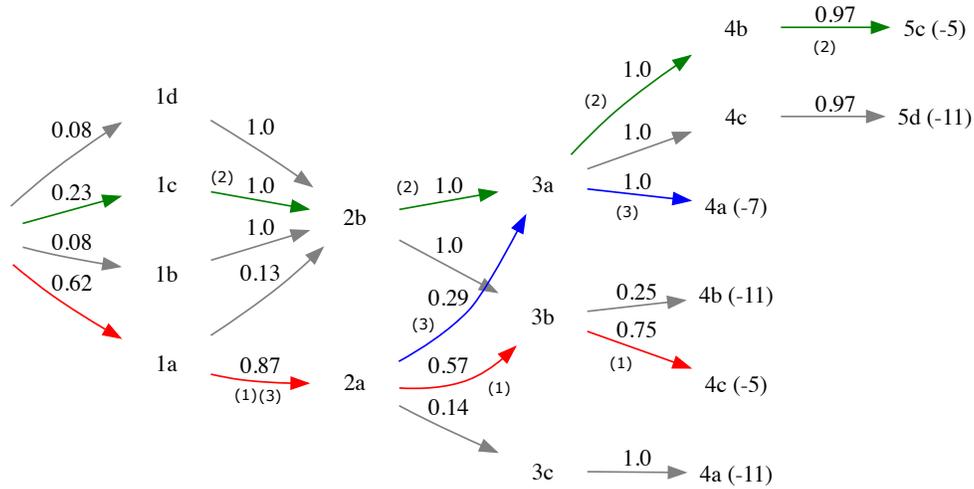

Figure 3 : *Task model* extracted at the end of the simulation

In conclusion, this simulation validates the construction of the predictive model in accordance with probability theory, for a stationary data regime and according to an incremental iterative approach. Furthermore, we have shown that it was possible to extract an easily interpretable *task model* from it, offering a fine representation of how the user employs their interface.

The following paragraph deals with the *sequential learning* capability of the ABIT-H algorithm. This property is generally associated with continuous learning. It is of crucial importance in the context of human-machine interaction. We show its qualitative aspects through a dynamic simulation.

## 5   SIMULATION OF SEQUENTIAL LEARNING

Unlike classical learning where all tasks are learned simultaneously from a balanced dataset, in *sequential learning* not all tasks are present in the data at the same time, as some will appear later, or even disappear after a certain period. The whole challenge is to have a predictive model capable of learning new tasks without having to replay past data, which implies keeping previous learning in memory. That said, the model must also be able to unlearn and forget habits that have become obsolete. In this new scheme, the relationship between human and machine is subject to constant adjustment on both sides, in a form of co-learning. Here we simulate a user's usage habits that evolve over time. The tasks considered correspond to navigation paths in a hierarchical menu characterized by a quasi-tree structure with 6 levels of depth, as defined below:

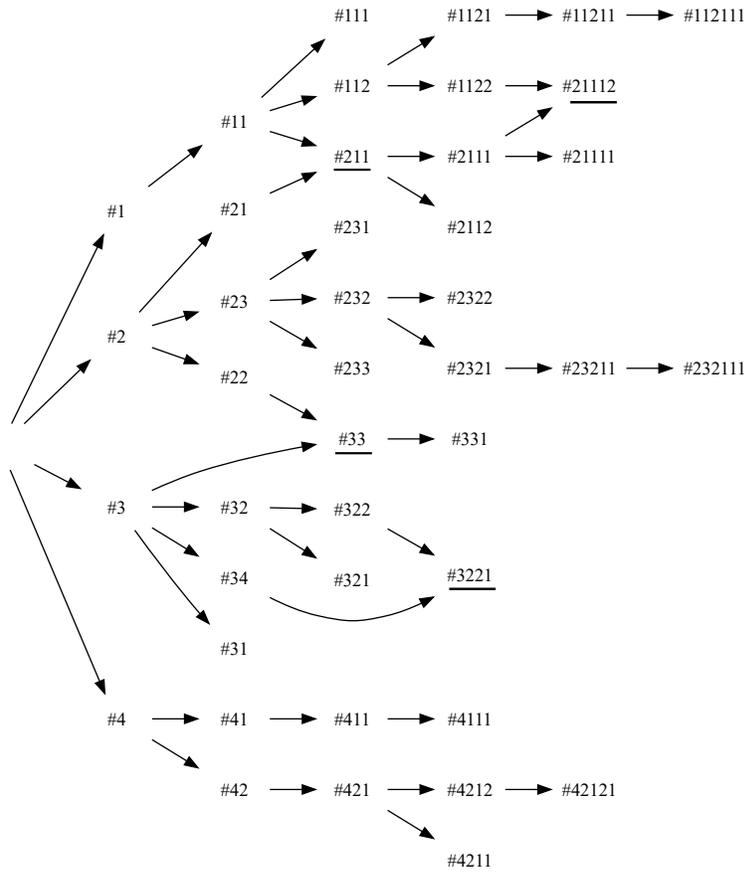

Figure 4 : Hierarchical menu structure

Some menus have two incoming paths and are underlined, such as menu #211. For this complex structure, a higher-order Markovian modeling is required to take into account dependency relationships between distant menus and thus be able to resolve prediction ambiguities of paths. We therefore keep the default configuration as it is associated with an optimal *Markov order*. The simulation consists of sequentially learning 4 groups of tasks. We selected 20 representative paths from the hierarchical menu as a whole. They are arbitrarily divided into 4 groups of 5 paths, as indicated in the table (Table 4):

Table 4 : The simulation paths divided into four groups

| Group N°. | path number | Paths |
| --- | --- | --- |
| **1** |  | #2 #21 #211 #2112 |
|  | (3) | #1 #11 #111 |
|  | (1) | #3 #33 #331 |
|  | (2) | #4 #42 #421 #4211 |
|  |  | #4 #42 #421 #4212 #42121 |

| 2 | (6) | #2 #21 #211 #2111 #21112 |
|---|---|---|
|   | (4) | #3 #34 #3221 |
|   |   | #2 #22 #33 #331 |
|   | (5) | #2 #23 #233 |
|   |   | #2 #23 #232 #2321 #23211 #232111 |
| 3 |   | #3 #32 #321 |
|   |   | #4 #41 #411 #4111 |
|   | (9) | #1 #11 #112 #1122 #21112 |
|   | (7) | #3 #31 |
|   | (8) | #2 #23 #232 #2322 |
| 4 |   | #1 #11 #112 #1121 #11211 #112111 |
|   |   | #3 #32 #322 #3221 |
|   | (11) | #2 #23 #231 |
|   |   | #2 #21 #211 #2111 #21111 |
|   | (10) | #1 #11 #211 #2112 |

Phase 1 consists of learning the paths of group 1 through random sampling of 50 sequences among the 5 paths (or tasks) of this group. Phase 2 proceeds in the same way by random sampling but this time among the 5 paths of group 2. The same applies to phases 3 and 4. Consequently, each learning phase focuses on new data related to new tasks. These 4 phases are performed one after the other, with the predictive model being continuously updated through incremental learning. In total, 200 data points are thus used to learn 20 tasks, each being more or less represented in the data. Indeed, this random sampling, based on a uniform law, naturally induces a large disparity in the occurrences of each navigation path, thus simulating usage habits of varying strength. At the end of each learning phase, the algorithm extracts the current *task model*, which we display below as results (see Figure 5).

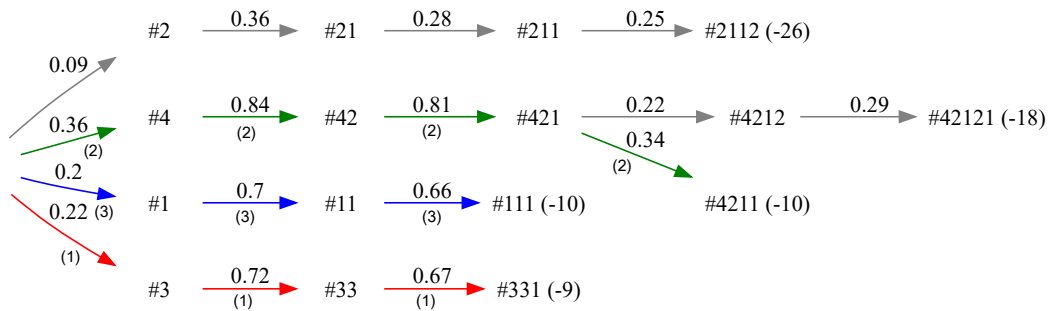

Figure 5 : *Task model* extracted after phase 1

At the end of phase 1, we observe that the 5 paths of group 1 are learned as expected. Their joint probability (measured in *decibans*) is indicated at the end of the path. Relatively strong *evidence* will reflect frequent habitual use, and conversely a weak *evidence*, an occasional use. Conditional probabilities are also indicated at the location of each transition between two menus, giving the information of the probability of the target menu given the previous menus. At this stage, the size of the predictive model is 120 parameters. We display below the 3 most probable paths identified in the figure by the numbers (1), (2) and (3), or associated with the colors: red, green and blue, namely (see Table 5 and Figure 5):

Table 5 : Ranked most probable paths (1), (2) and (3) after phase 1

| Rank | Path number | Path sequence | Evidence |
|---|---|---|---|
| 1 | (1) | #3(0.22) #33(0.72) #331(0.67) | −9 dB |
| 2 | (2) | #4(0.36) #42(0.84) #421(0.81) #4211(0.34) | −10 dB |
| 3 | (3) | #1(0.2) #11(0.7) #111(0.66) | −10 dB |

At the end of phase 2, the model size is 237 parameters. We observe (see Figure 6) that the *task model* now contains the structural and statistical information of the 10 reference sequences from groups 1 and 2.

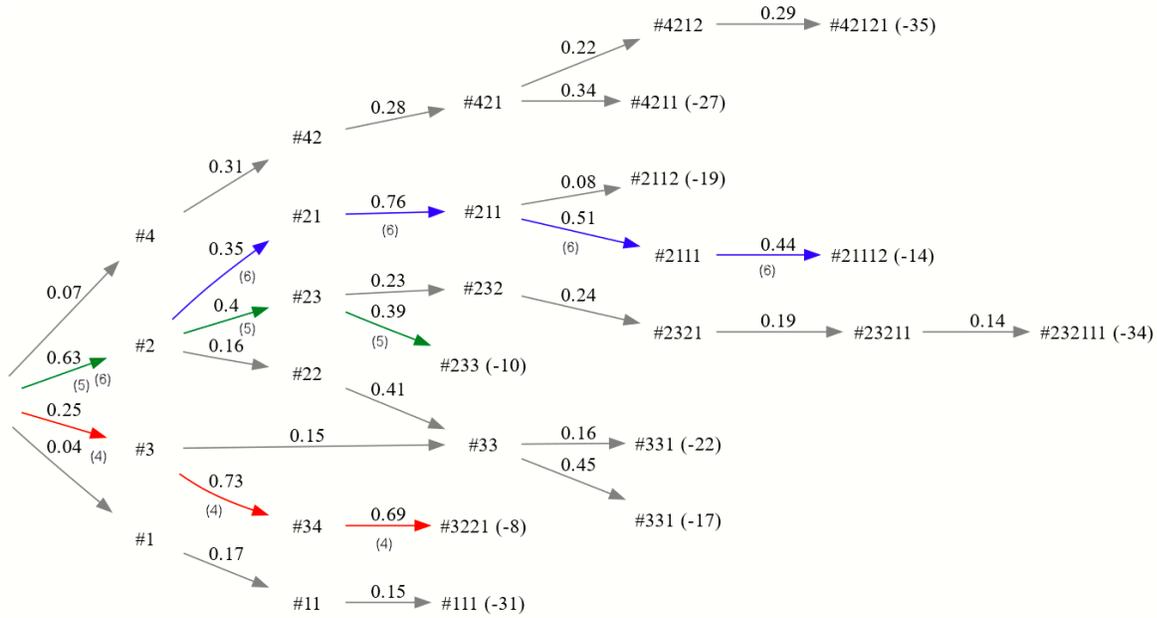

Figure 6 : Task model extracted after phase 2

The 3 most probable paths are then (see Table 6 and Figure 6):

Table 6 : Ranked most probable paths (4), (5) and (6) after phase 2

| Rank | Path number | Path sequence | Evidence |
|---|---|---|---|
| 1 | (4) | #3(0.25) #34(0.73) #3221(0.69) | −8 dB |
| 2 | (5) | #2(0.63) #23(0.4) #233(0.39) | −10 dB |
| 3 | (6) | #2(0.63) #21(0.35) #211(0.76) #2111(0.51) #21112(0.44) | −14 dB |

They are distinct from the previous ones and belong to group 2. This result is explained by the fact that they correspond to more recent habits, knowing that old habits (those of group 1) see their *evidence* decrease. For example, the *evidence* of the sequence "#3 #33 #331" (path number (1) in Figure 5) goes from −9 dB to −17 dB, reflecting the mechanism of progressive forgetting in the absence of occurrence. Note that this simulation uses a very restricted number of training data, namely 50 examples per learning phase (for an analysis window of size 32), which does not allow the algorithm to converge to a stationary state, hence, as expected, low probabilities at the output (negative *evidence*). Recall that Bayesian probabilities account for usage frequencies but also for the amount of statistical information. Finally, like the predictive model, the *task model* is estimated "at its best" given past experience.

At the end of phase 3, the size of the predictive model is 311 parameters. The knowledge model now reflects the 15 tasks learned during phases 1, 2, and 3 (see Figure 7).

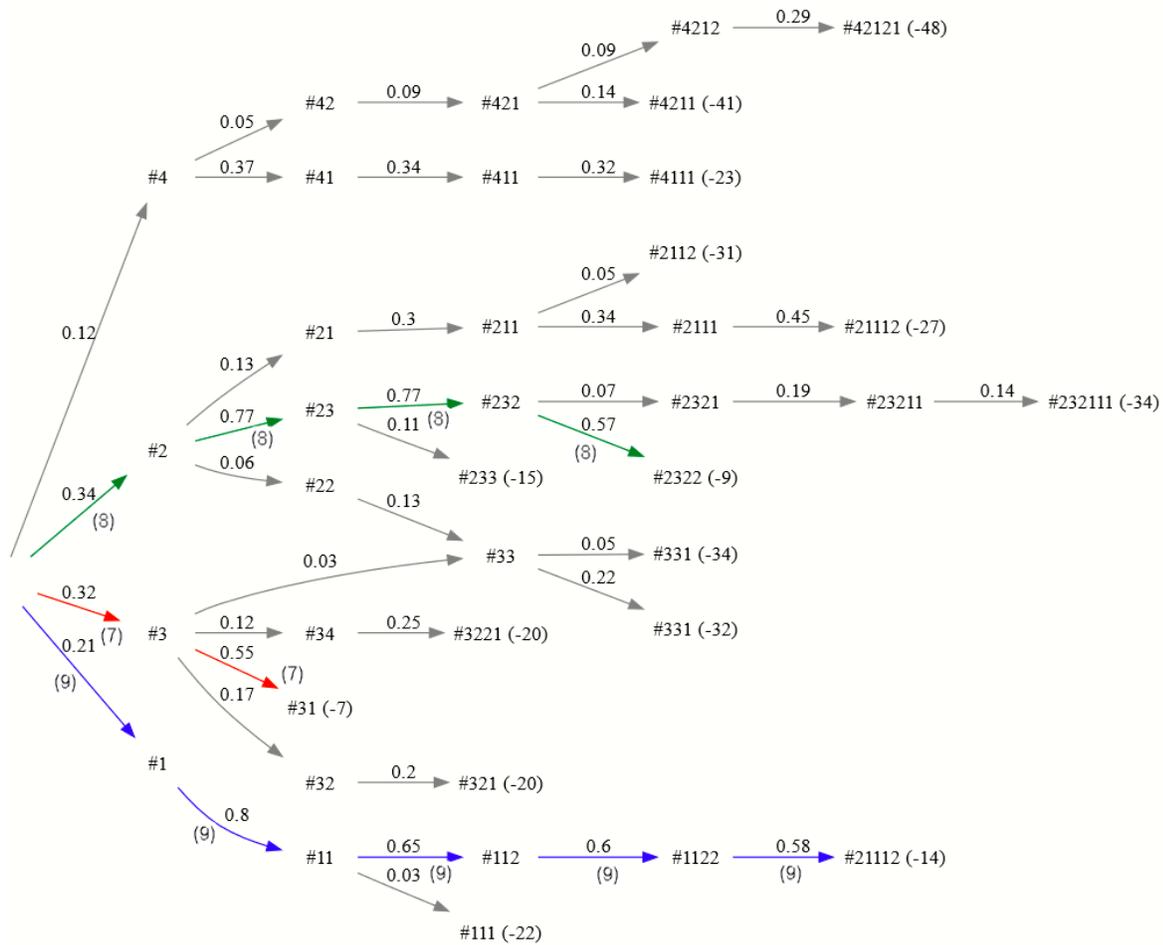

Figure 7 : *Task model* extracted after phase 3

The 3 most probable paths are (see Table 7):

Table 7 : Ranked most probable paths (7), (8) and (9) after phase 3

| Rank | Path number | Path sequence | Evidence |
|---|---|---|---|
| 1 | (7) | #3(0.32) #31(0.55) | −7 dB |
| 2 | (8) | #2(0.34) #23(0.77) #232(0.77) #2322(0.57) | −9 dB |
| 3 | (9) | #1(0.21) #11(0.8) #112(0.65) #1122(0.6) #21112(0.58) | −14 dB |

According to Table 4, these paths indeed belong to group 3 containing the most recent habits at this stage of the simulation. We observe that the probability of the sequence "#3 #33 #331" (path number (1) in Figure 5) from group 1 has decreased further, going from −17 dB to −32 dB. Indeed, this sequence having had no new occurrences since phase 1, it tends to gradually disappear from the model.

Finally, at the end of phase 4, all 20 paths have been learned, and the knowledge graph faithfully reflects the complete structure of the hierarchical menu (see Figure 8). The size of the predictive model is 390 parameters. The ranking of the top 3 paths is as follows (see Table 8):

Table 8 : Ranked most probable paths (10), (11) and (7) after phase 4

| Rank | Path number | Path sequence | Evidence |
|---|---|---|---|
| 1 | (10) | #1(0.46) #11(0.98) #211(0.5) #2112(0.71) | −7 dB |
| 2 | (11) | #2(0.34) #23(0.61) #231(0.55) | −9 dB |
| 3 | (7) | #3(0.17) #31(0.16) | −15 dB |

The two most probable paths belong to group 4, but we note that the third path (path number (7)) belongs to group 3. This is explained by the fact that the latter was associated with a relatively strong habit developed during phase 3. Its *evidence* then decreased as expected during phase 4, going from −7 dB to −15 dB, while remaining high enough to be better ranked than paths from group 4.

In summary, the 4 groups of tasks have been learned sequentially without catastrophic forgetting. The model has kept old tasks in memory while progressively revising their probability downward, accounting for their obsolescence. As for new tasks, they were each time quickly considered by the probabilistic model. The evolution of probabilities has therefore proven to be consistent with the order of appearance of data and tasks. At all stages of learning, the *task model* has faithfully highlighted the logical sequence of actions, this for an optimal *Markov order*. Note that we have replayed this simulation many times by changing the order of appearance of tasks, without any notable differences in the results.

We conclude that this simulation validates the feasibility of a *sequential learning* that is both robust and continuous based on purely Bayesian modeling, thus demonstrating the ability of this approach to handle a dynamic environment. Note that until now we have generated the *task model* from the root (i.e., with an empty prompt). However, it is possible to generate it from any point in the hierarchical menu, typically based on the user's current path.

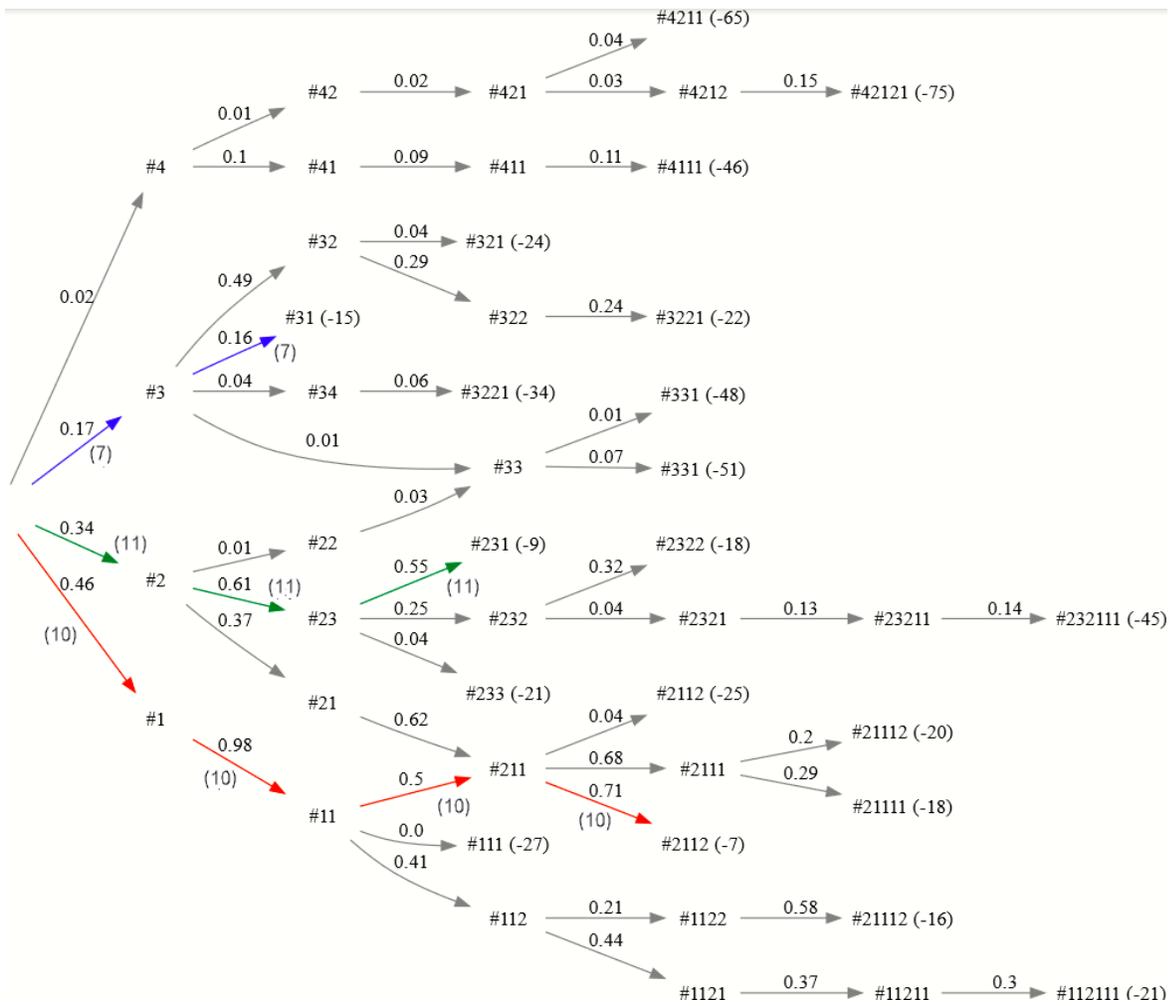

Figure 8 : Task model extracted after phase 4

For example, if the user has already validated menus #1 and #11, then the prompt "#1 #11" induces the following subgraph (see Figure 9):

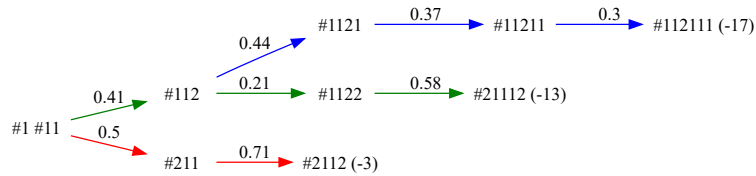

Figure 9 : Task model generated from the prompt "#1 #11"

And if the user then validates menu #112, the new prompt induces the subgraph (see Figure 10):

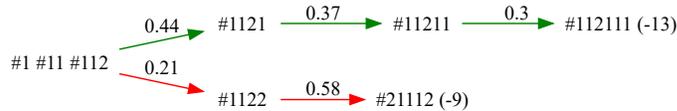

Figure 10 : Task model generated from the prompt "#1 #11 #112"

The algorithm can thus recalculate the *task model* at each step of the user's journey, redefining the field of their predictable actions and updating the probabilities of goals.

In summary, this adaptive Bayesian approach provides a particularly rich model of activity and behavior highlighting the user's usage habits as well as the logical structure of the interface. From this model, it becomes possible to take them into account to trigger interface adaptations and thus optimize interaction. We envision a system to help the user better navigate their interface, for example by highlighting their usual paths, thus reducing their mental load. In the case of marked habits, the system will also be able to propose action shortcuts (or macro-commands) allowing the user to reach their goals more quickly. Finally, in the case of a high level of confidence in predictions, depending on the nature of the task (its criticality) and the benefit/cost ratio, the system will be able to automate all or part of the task. The precise definition of these dynamic adaptations is a matter of UX design choices.

## 6 CONCLUSION

In this article, we have presented an adaptive Bayesian inference technique for modeling user habits in a "small data" approach. This approach focuses on specific data from the current user's activity, in contrast to "big data" techniques that deal with complex data from multiple sources. Its objective is to facilitate user interface personalization by allowing real-time adjustment of service usage.

We have established the theoretical framework for this technique and conducted a numerical simulation demonstrating that such a probabilistic approach is capable of optimally modeling and predicting a sequence of actions. This rigorous modeling, achievable with a limited number of variables, can produce reliable confidence measures on output predictions, which is crucial in a UX context where the system may make autonomous decisions leading to interface modifications. We have shown that excessive simplification of the predictive model, obtained by choosing a *Markov order* that is too small, led to erroneous predictions. This highlights the importance of capturing all structural dependency relationships between the data, thus ensuring robust inference.

Because humans are constantly changing beings, we focused on continuous and sequential learning in a dynamic environment. In this context, Bayesian inference gives the model the ability to rationally revise its beliefs with each new data point in the presence of uncertainty. We implemented this in an online incremental learning mode using the ABIT-H algorithm whose principles have been presented. We tested it through a simulation of *sequential learning* of navigation tasks in a hierarchical menu, thus demonstrating its ability to quickly learn new tasks robustly while maintaining its past knowledge.

Finally, based on this approach, we have shown that it is possible to extract in real time a knowledge model of the user's interactions with their machine, highlighting their usage statistics and reflecting the logical structure of the application. This *task model*, both quantitative and qualitative, allows understanding how the user uses their interface: what their habits are, what their level of mastery is, what their potential obstacles are. It opens the way to various interface adaptation strategies to help the user accomplish their tasks while reducing their cognitive load. The next step will be to integrate this technology into an existing mobile application and experiment with different adaptivity paradigms, particularly guidance and action shortcuts.